%% file: main.tex
\documentclass[10pt,twocolumn,letterpaper]{article}

\usepackage[pagenumbers]{cvpr} 

\input{preamble}
\definecolor{cvprblue}{rgb}{0.21,0.49,0.74}
\usepackage[pagebackref,breaklinks,colorlinks,allcolors=cvprblue]{hyperref}

\def\paperID{(*****)} 
\def\confName{CVPR}
\def\confYear{2026}

\title{Rethinking Infrastructure Inspection as Image Difference Classification: A Traffic Sign Case Study}

\author{Ching Yau Fergus Mok\\
University of Cambridge\\
Cambridge, UK\\
{\tt\small cym23@cam.ac.uk}
\and
Lavindra de Silva\\
University of Cambridge\\
Cambridge, UK\\
{\tt\small lpd25@cam.ac.uk}
\and
Varun Kumar Reja\\
IIT Bombay\\
Mumbai, India\\
{\tt\small varunreja@iitb.ac.in}
\and
Ioannis Brilakis\\
University of Cambridge\\
Cambridge, UK\\
{\tt\small ib340@cam.ac.uk}
}

\begin{document}
\maketitle
\input{sec/0_abstract}    
\input{sec/1_intro}
\input{sec/2_related_work}

\input{sec/3_methodology}
\input{sec/4_results_and_discussion}

\input{sec/5_discussion}
\input{sec/6_ablations}
\input{sec/7_conclusion}
{
    \small
    \bibliographystyle{ieeenat_fullname}
    \bibliography{main}
}


\end{document}

%% file: sec/0_abstract.tex
\begin{abstract}
Digital twins (DTs) allow the digitalization of road infrastructure inspection, though this is hindered by limited annotated data. This work exploits the relational nature of continuous asset condition monitoring to reformulate image-based defect detection as image difference classification (IDC) to reduce data reliance. This was evaluated in a case study on low-resource traffic sign inspection with different IDC classifiers using a newly-curated, high quality dataset. Results indicate that the instruction-based classifier outperforms encoder-based ones and gains from comparison with reference images. This shows that IDC can be an effective task modeling for tackling data constraints in infrastructure inspection and DT asset condition updating.
\end{abstract}

%% file: sec/1_intro.tex
\section{Introduction}
\label{sec:intro}

Maintaining road infrastructure requires manual visual condition inspection. The proposal of a Road Digital Twin (DT) \cite{digital_twin_nh}, coinciding with increasing road imagery capturing from smart vehicles, presents an opportunity to digitalize this process. Particularly, the Road DT framework tracks the evolving condition of road assets through successive updating steps each involving classifying an image into different condition labels specified by asset-specific data requirements \cite{reja2026digitaltwin}. 

Annotated data for road maintenance is scarce like in many Architecture, Engineering, and Construction (AEC) fields. To reduce data dependency, this work exploits the relational nature of the DT updating step by rethinking defect identification as an image difference classification (IDC) task where an inspection image is compared against a reference image of the same asset taken previously to produce a grounded output. Asset images over time already exist in real road asset management systems (AMSs), and this task formulation uses them as references without needing additional annotation. Due to the wide range of assets, this work focuses on traffic signs as they are both critical to road safety and contain a large number of failure modes.

The contributions of this paper are: (1) Curating a high quality and versatile traffic sign imagery dataset with condition annotation. (2) Experimenting with IDC pipelines on traffic sign inspection tasks and quantifying the effects of adding a reference image in reducing data-dependency in low-resource settings.

%% file: sec/2_related_work.tex
\section{Related Work}
\label{sec:related work}

Vision models are increasingly capable due to large-scale pre-training which uses large data sources to inject general knowledge to models. They have been used in various IDC and traffic sign maintenance related work.

\subsection{Vision models and IDC}
\label{subsec:vision language models}
A notable pre-trained family is CLIP \cite{clip_op} which performs image-text and image-image comparisons with embeddings similarly to Sentence Transformers for text \cite{sbert_op}. State-of-the-art vision-encoders include DINOv3 \cite{dino_v3_op}, SigLIP2 \cite{siglip_2_op}, and MetaCLIP2 \cite{clip_2_op}. Extending from Large Language Models, increasingly popular generative vision language models (VLMs) allow for instruction-tuning. The most sophisticated close-source models are unsuitable for privacy-sensitive AEC tasks like road maintenance which use government data, so this work only considers open-sourced families like Qwen VL \cite{qwen3_vl_op} and Gemma \cite{gemma_3_op}.

Vision models have been used for image difference tasks. \cite{Dou_2025_ICCV} enforced orthogonality to improve IDC for subtle changes. \cite{sam2025} improved difference reasoning with CLIP by aligning image embedding differences with texts. \cite{guo-etal-2022-clip4idc} adapted CLIP for difference captioning by designing an image-image-text fusion architecture, while \cite{onediff} introduced a difference captioning VLM with learned delta tokens.

\subsection{Computer vision in traffic sign maintenance}
\label{subsec:computer vision in traffic sign maintenance}

Traffic sign imagery datasets with condition annotations are scarce. \cite{sandhu2023} annotated the Mapillary sign dataset \cite{mapillary_dataset} with five condition labels and augmented them to form 20,000 examples while \cite{nagy2021} labeled 4,000 sign images from dashboard cameras into four conditions. \cite{rados2022} synthetically added graffiti to signs from the GTSDB dataset \cite{GTSDB}. \cite{chen2023mflyolo} annotated signs into binary condition labels and augmented them to form 2164 damaged and 4330 undamaged signs. For task modeling, \cite{chen2023mflyolo} created a binary classifier for defect presence, \cite{nagy2021} classified for defect types, while \cite{ersu2024} and \cite{rados2022} classified for severity modeled by occlusion amount.

To the best of the authors' knowledge, existing work misaligns with real sign inspection needs and there remain large gaps in research. Firstly for datasets, many only contain close-ups of the sign face (without post or background) \cite{sandhu2023}\cite{nagy2021} or very coarse condition labeling (binary or one defect) \cite{rados2022}\cite{chen2023mflyolo}\cite{ersu2024}, and none allows for condition tracking over time. Secondly for task modeling, none treats defect identification as a multi-class multi-label task (different defects often occur simultaneously), none leverages multiple images for detection (simulating Road DT updating and condition tracking), with all using only single-images and most relying solely on augmentation to tackle data scarcity \cite{sandhu2023}\cite{rados2022}.

%% file: sec/3_methodology.tex
\section{Methodology}
\label{sec:methodology}

To cover the gaps, this section introduces a new dataset and discusses the objectives and experiments performed.

\subsection{Dataset}
\label{subsec:dataset creation}
Traffic sign images were collected from a real road AMS used by National Highways, the UK highway authority. This dataset fills the gap by providing: (1) Real inspection imagery showing all parts of a sign (sign face, post, and background) to capture more failure modes. (2) Two images per unique sign (undamaged reference + inspection) to track conditions. (3) Fine-grained multi-label condition annotation to match real maintenance needs.

The dataset (available \href{https://huggingface.co/datasets/PastaEvangelists/UK_Traffic_Sign_Inspection_Dataset}{here}\footnote{Provisional release, full dataset with additional annotations like sign attributes will be released later.}) contains 970 image pairs along with annotations into nine condition categories (\textit{sign post tilted or fallen, sign post deformed, sign post rotten, sign face detached or insecure, sign face deformed, sign face aged, sign face vandalized, sign face dirty, and a no defect flag}), extending from at most five in other work \cite{sandhu2023}. Note that significant class imbalance exists between categories, with counts from 20 (sign face vandalized) to 184 (sign face detached or insecure). However, the numbers of damaged and undamaged signs are balanced ($\sim$4:6).

\subsection{Objective and experiments}
\label{subsec:vision model experimentation}
The objective of this work is to investigate whether reference images can improve traffic sign defect classification performance through image-image comparison and reduce data-dependency in low-resource settings. The authors hypothesize that asset images over time can be used as references to supplement training data.

Two tasks were studied: binary defect presence detection and multi-class multi-label defect classification. The former is relevant to filtering systems for prioritizing images for manual review, while the latter can be directly applicable to Road DT asset condition updating. Both tasks were modeled as a combined multi-class multi-label IDC problem with an additional flag for defect presence. Data constraints were simulated by fine-tuning classifiers using different numbers of examples per class (shot). The performance of IDC pipelines was then compared with the best performing equivalent single-image pipelines that do not use reference images.

Experiments used an RTX 4080 with 16GB VRAM, using libraries including Transformers, Pytorch, and Unsloth. All training-set reference images were used during fine-tuning as pseudo ``no defect" examples to ensure models see the same images irrespective of whether classification uses references. Every experiment was repeated with five different data splits to mitigate few-shot learning instability (similarly to \cite{setfit}), with averaged results reported.

Classifier pipelines are divided into encoder-based and instruction-based. They are described in detail below.

\subsubsection{Encoder Based Pipelines}
\label{susubsec:vision encoders}

Four encoder-based pipelines were proposed. Two IDC classifiers are purely vision-based: one simply uses a linear classifier to fuse the two concatenated image embeddings while the other adds an image-image cross attention layer (modeled after the cross attention block in the original transformer) prior to the linear layer to allow explicit interaction. Two more IDC classifiers also use textual class descriptions: one adds an image-image-text cross attention layer prior to the linear classifier, while the other uses similarity between image embedding difference and class texts as classification logits (similarly to \cite{sam2025}). The equivalent single-image pipeline for purely vision-based IDC classifiers was chosen as a linear classifier on the single-image embedding, while those for the text-enhanced classifiers were chosen as image-text cross attention prior to the linear classifier and single-image embedding similarity with class texts. MetaCLIP2 was chosen as the backbone as it outperformed other models tested (e.g.\ SigLIP2).

\subsubsection{Instruction Based Pipeline}
\label{subsubsec:instruction-based vlms}

Instruction-based VLMs have shown strong performance in comparative reasoning \cite{sam2025}. The inspection and reference images are inserted into the user prompt where it is instructed to use the latter as a classification guide. Textual class descriptions are provided in the system prompt. Structured classification outputs are collected from model response in json. Primary investigations focused on fine-tuning the model based on example interactions. Qwen3 8B was chosen as the model as it outperformed other models tested (e.g.\ Gemma 3).

%% file: sec/4_results_and_discussion.tex
\section{Results}
\label{sec:results}
IDC results of the two experimented tasks are explored in this section, with metric values reported being the averages over five data splits to mitigate few-shot instability.

\subsection{Defect presence detection performance}
\label{subsec:defect presence detection performance}
Defect presence detection was modeled as a binary classification task with a dedicated defect presence flag. \cref{tab:no defect f1} shows the IDC $f1$ scores for this task along with the improvements over the best equivalent single-image pipelines that do not use reference images. Most pipelines perform this task well, with many attaining $\sim$0.9 $f1$ even when trained with very few examples per class (shot). This is expected as detecting whether a sign contains a defect is a simple binary task with only two possible outcomes. With the exception of embedding difference similarity, all encoder-based pipelines attain comparable results with none having a significant and consistent lead. However, the instruction-based pipeline consistently outperforms all encoder-based ones, with $f1$ reaching above 0.9 at just 1-shot.

\begin{table*}
  \caption{Binary defect presence detection $f1$ of IDC pipelines along with improvements over the best equivalent single-image pipelines ( $f1_{\pm \text{improvement}}$, $+$ means IDC is better). Shots represent minimum number of examples per class. LC stands for Linear Classifier. First four rows are encoder-based classifiers with MetaCLIP2 (2B), final row is an instruction-based classifier with Qwen3-VL-8B.}
  \label{tab:no defect f1}
  \centering
  \begin{tabular}{@{}lccccc@{}}
    \toprule
    IDC $_{\textit{Best Equivalent Single-Image}}$ & 0-shot & 1-shot & 2-shot & 4-shot & 8-shot \\
    \midrule
    LC $_{\textit{LC}}$ & N/A & $0.679_{-0.142}$ & $0.767_{-0.102}$ & $0.870_{-0.020}$ & $0.884_{+0.013}$ \\
    Attn.\ (im-im) + LC $_{\textit{LC}}$ & N/A & $0.800_{-0.020}$ & $0.864_{-0.005}$ & $0.877_{-0.013}$ & $0.897_{+0.026}$ \\
    Attn.\ (im-txt\&im) + LC $_{\textit{Attn. (im-txt) + LC}}$ & N/A & $0.790_{-0.035}$ & $0.851_{+0.007}$ & $0.885_{+0.007}$ & $0.891_{-0.007}$\\
    Emb.\ Diff.\ Sim.\ $_{\textit{Attn. (im-txt) + LC}}$ & N/A & $0.589_{-0.236}$ & $0.625_{-0.218}$ & $0.681_{-0.198}$ & $0.687_{-0.211}$\\
    Instruction $_{\textit{Instruction}}$ & $0.007_{-0.222}$ & $0.905_{+0.009}$ & $0.935_{+0.017}$ & $0.938_{+0.012}$ & $0.940_{+0.031}$\\
    \bottomrule
  \end{tabular}
\end{table*}

Encoder-based models (first four rows) do not appear to benefit from the addition of a reference image, with IDC pipelines shifting between being better and worse than the single-image pipelines evidenced by the constantly changing signs. The instruction-based IDC pipeline, however, is able to consistently improve over the single-image pipeline, with every training scenario showing improvements between 0.009 and 0.031. Interestingly though, the untrained instruction-based IDC pipeline classifies almost all images as having a defect, much more so than when not using reference images, resulting in an almost 0 score. This suggests that fine‑tuning on just one example per class already effectively “calibrates” the model, showing what a defect is and how to use the reference image.


\subsection{Defect classification performance}
\label{subsec:defect classification performance}

The second task goes further than simply detecting defect presence by classifying the defect type(s), modeling this as a multi-class multi-label classification task. \cref{tab:macro f1} shows the $macro$ $f1$ scores (averaged over classes) of different IDC pipelines along with the improvements over the best equivalent single-image pipelines that do not use reference images. Scores for this task are lower than the first task due to the increased complexity, with none of the encoder-based pipelines (first four rows) exceeding 0.5 and the instruction-based one achieving just over 0.6. As before, the instruction-based pipeline has a consistent lead over all encoder-based ones, with its 1-shot score exceeding even the 8-shot scores of encoder-based pipelines.

\begin{table*}
  \caption{Multi-class multi-label defect classification $macro$ $f1$ of IDC pipelines along with improvements over the best equivalent single-image pipelines ( $f1_{\pm \text{improvement}}$, $+$ means IDC is better). Shots represent minimum number of examples per class. LC stands for Linear Classifier. First four rows are encoder-based classifiers with MetaCLIP2 (2B), final row is an instruction-based classifier with Qwen3-VL-8B.}
  \label{tab:macro f1}
  \centering
  \begin{tabular}{@{}lccccc@{}}
    \toprule
    IDC $_{\textit{Best Equivalent Single-Image}}$ & 0-shot & 1-shot & 2-shot & 4-shot & 8-shot \\
    \midrule
    LC $_{\textit{LC}}$ & N/A & $0.266_{-0.080}$ & $0.327_{-0.033}$ & $0.402_{-0.023}$ & $0.449_{+0.005}$ \\
    Attn.\ (im-im) + LC $_{\textit{LC}}$ & N/A & $0.321_{-0.025}$ & $0.371_{+0.011}$ & $0.388_{-0.037}$ & $0.470_{+0.026}$ \\
    Attn.\ (im-txt\&im) + LC $_{\textit{Attn. (im-txt) + LC}}$ & N/A & $0.164_{-0.071}$ & $0.324_{+0.008}$ & $0.410_{-0.002}$ & $0.479_{-0.010}$\\
    Emb.\ Diff.\ Sim.\ $_{\textit{Attn. (im-txt) + LC}}$ & N/A & $0.206_{-0.029}$ & $0.216_{-0.099}$ & $0.214_{-0.197}$ & $0.219_{-0.269}$\\
    Instruction $_{\textit{Instruction}}$ & $0.346_{-0.007}$ & $0.544_{+0.033}$ & $0.536_{+0.012}$ & $0.568_{+0.008}$ & $0.601_{+0.038}$\\
    \bottomrule
  \end{tabular}
\end{table*}


The addition of reference images again does not improve the encoder-based scores, while the instruction-based pipeline benefits from it at every shot, with gains ranging from 0.008 to 0.038. The only exception is at 0-shot where providing the reference image to the untrained model degrades performance, again demonstrating that ``calibration" through fine-tuning is needed to take advantage of the extra input. Statistical significance of the gains from reference images in the fine-tuned instruction-based pipeline was evaluated through a two-sided paired t-test across data splits, with statistics shown in \cref{tab:instruc_micro_f1_ttest}. 1 and 8-shots see statistically significant improvements with very low p-values while 2 and 4-shots do not, despite all settings still showing a consistently positive trend. This is likely due to instability from training with very few examples in few-shot learning, with the standard deviation being especially big at 2-shot.

\begin{table}
  \caption{Statistics of improvements in defect classification $macro$ $f1$ of instruction-based IDC over the single-image pipeline.}
  \label{tab:instruc_micro_f1_ttest}
  \centering
  \begin{tabular}{@{}lccccc@{}}
    \toprule
    Statistics & 1-shot & 2-shot & 4-shot & 8-shot\\
    \midrule
    standard deviation & $0.012$ & $0.035$ & $0.015$ & $0.014$\\
    p-value & $0.006$ & $0.543$ & $0.366$ & $0.006$\\
    \bottomrule
  \end{tabular}
\end{table}

%% file: sec/5_Discussion.tex
\section{Discussion}
\label{subsec:Discussion}
It is important to note that all pipelines (IDC and single-image) have access to the same images during fine-tuning, as reference images are still used as pseudo ``no defect" examples irrespective of whether they are used for classification. Therefore, any improvement is a result of the additional context of optimal asset states provided by reference images and not of an effective increase in data availability.

That being said, two main observations can be made: Firstly, 
the instruction-based classifier outperforms all encoder-based ones and is able to leverage reference images for performance gains. A potential reason for this gap can be the different number of model parameters, which will be explored in the Ablations in \cref{subsec:ablations}. Secondly, models require ``calibration" before being able to make use of the additional reference input. This can be done by fine-tuning with as few as one example per class, which raised $f1$ for binary defect presence detection from 0.07 to 0.905.

With these observations, IDC can be an effective task modeling for continuous infrastructure inspection and DT condition updating. Reference images can improve detection results and supplement training data in low-resource settings. However, this is limited to only instruction-based classifiers and requires a small ``calibrating" fine-tuning set. A drawback of this is that instability from few-shot learning can create fluctuating outputs which can affect reliability.

%% file: sec/6_ablations.tex
\section{Ablations}
\label{subsec:ablations}
The following ablation studies were carried out.

\subsection{Bigger encoder backbone}
\label{subsubsec:bigger encoder backbone}

To investigate whether using a bigger vision-encoder comparable to the size of the instruction-based backbone can bridge the gap in performance between the encoder-based and instruction-based pipelines, the 7B DINOv3 was tested in addition to the 2B MetaCLIP2. \cref{tab:clip vs dino f1} shows $macro$ $f1$ scores of the two vision-only pipelines with the two backbones. MetaCLIP2 significantly outperforms DINOv3 in every training shot despite being smaller in size. Pure number of model parameters does not seem to be the sole determinant of the difference in performance between encoder-based and instruction-based pipelines.

\begin{table}
  \caption{Multi-class multi-label defect classification $macro$ $f1$ of IDC pipelines for MetaCLIP2 (2B) and DINOv3 (7B).}
  \label{tab:clip vs dino f1}
  \centering
  \begin{tabular}{@{}lcccc@{}}
    \toprule
    & \multicolumn{2}{c}{CLIP2 (2B)} & \multicolumn{2}{c}{DINOv3 (7B)} \\
    Pipeline & 1-shot & 8-shot & 1-shot & 8-shot\\
    \midrule
    LC & $0.266$ & $0.449$ & $0.125$ & $0.360$\\
    Attn.\ (im-im) + LC & $0.321$ & $0.470$ & $0.168$ & $0.327$\\
    \bottomrule
  \end{tabular}
\end{table}

\subsection{Pseudo defects}
\label{subsubsec:pseudo defects}

In the main setup, reference images were used as ``no defect" pseudo defect images to keep training data identical between pipelines to avoid unintended gains from an effective increase in data availability when using reference images. Experiments without pseudo defect images were performed for the instruction-based pipeline to understand their effects on performance. Results show that their use improves performance across all training scenarios (e.g.\ +0.016 for 8-shot IDC), and that the IDC pipeline maintains its lead over the single-image pipeline even without pseudo defects (e.g.\ +0.053 for 8-shot).

%% file: sec/7_conclusion.tex
\section{Conclusion}
\label{sec:conclusion}

The Road DT aims to digitalize road asset condition monitoring. This work exploits the relational nature of continuous asset tracking to reduce data dependency by rethinking defect identification as IDC. This uses already-existing asset images over time as references for grounded classification through image comparison. Using a newly curated dataset of real inspection images, this task formulation was tested in a case study on low-resource traffic sign inspection. Results show that the instruction-based classifier consistently outperforms encoder-based ones, and that with a small fine-tuning dataset of as few as one example per class, it is able to leverage the additional reference inputs for performance gains. Therefore, IDC can be an effective task modeling for reducing data reliance in continuous infrastructure inspection and DT condition updating.

Future work includes testing larger datasets to eliminate few-shot fluctuations and experimenting with using more than one reference images per sign.